\documentclass[conference]{IEEEtran}
\IEEEoverridecommandlockouts
\usepackage{cite}
\usepackage{amsmath,amssymb,amsfonts}
\usepackage{algorithmic}
\usepackage{graphicx}
\usepackage{textcomp}
\usepackage{xcolor}
\def\BibTeX{{\rm B\kern-.05em{\sc i\kern-.025em b}\kern-.08em
    T\kern-.1667em\lower.7ex\hbox{E}\kern-.125emX}}
\begin{document}

\title{Cannabis Seed Variant Detection using Faster R-CNN\\
}

\author{\IEEEauthorblockN{Toqi Tahamid Sarker}
\IEEEauthorblockA{\textit{School of Computing} \\
\textit{Southern Illinois University}\\
Carbondale, USA 62901\\
toqitahamid.sarker@siu.edu}
\and
\IEEEauthorblockN{Taminul Islam}
\IEEEauthorblockA{\textit{School of Computing} \\
\textit{Southern Illinois University}\\
Carbondale, USA 62901\\
taminul.islam@siu.edu}
\and
\IEEEauthorblockN{Khaled R Ahmed}
\IEEEauthorblockA{\textit{School of Computing} \\
\textit{Southern Illinois University}\\
Carbondale, USA 62901\\
khaled.ahmed@siu.edu}
}

\maketitle

\begin{abstract}
Analyzing and detecting cannabis seed variants is crucial for the agriculture industry. It enables precision breeding, allowing cultivators to selectively enhance desirable traits. Accurate identification of seed variants also ensures regulatory compliance, facilitating the cultivation of specific cannabis strains with defined characteristics, ultimately improving agricultural productivity and meeting diverse market demands. This paper presents a study on cannabis seed variant detection by employing a state-of-the-art object detection model Faster R-CNN. This study implemented the model on a locally sourced cannabis seed dataset in Thailand, comprising 17 distinct classes. We evaluate six Faster R-CNN models by comparing performance on various metrics and achieving a mAP score of 94.08\% and an F1 score of 95.66\%. This paper presents the first known application of deep neural network object detection models to the novel task of visually identifying cannabis seed types.
\end{abstract}


\section{Introduction}
Seed detection is vital in agriculture and horticulture to ensure quality, purity, and genetic diversity. In cannabis cultivation, it is particularly important due to the unique characteristics and regulatory restrictions imposed by different countries with this plant. The classification of cannabis into distinct categories, such as marijuana or industrial hemp, is primarily dependent on the threshold level of THC, the psychoactive compound found in the plant. Although 1\% THC is typically considered enough to cause intoxication, many jurisdictions establish the legal distinction between marijuana and hemp at the 0.3\% THC threshold \cite{Small1976-hv}. The cultivation of hemp is typically permitted only if it contains less THC than this defined threshold. Therefore, accurately identifying cannabis seeds with high or low THC content is essential for regulatory compliance and distinguishing between marijuana and industrial hemp \cite{Cherney2016-ae}. 

Moreover, in recent years, Cannabis has seen legalization in many countries in the world. The 2014 Farm Bill \cite{noauthor_undated-pk} legalized hemp pilot programs in the United States for research purposes, enabling the development of a domestic hemp industry after decades of prohibition. In the years since this policy change, it has proven economically beneficial for the United States. As seen in the trade data, US imports of hemp seeds, fibers, oils, and other ingredients have steadily risen, totaling \$67.3 million in 2017 \cite{Congressional_Service2018-ox}. Such legalization efforts have suddenly increased the demand for Cannabis for various purposes, such as recreational, medicinal, and industrial uses. In the hemp industry, beyond its medicinal value, hemp plays a pivotal role as a crucial resource for the manufacturing of ropes, textiles, and paper \cite{Cherney2016-ae}. The presence of undesired seeds can lead to cross-pollination, resulting in undesirable traits in the crop. Thus, identifying and detecting cannabis seeds with different variants ensures uniform product quality and a profitable yield.

Previous studies have applied a range of artificial intelligence methods for different applications in cannabis agriculture. For instance, Sieracka et al. \cite{Sieracka2023-dq} used artificial neural networks to forecast industrial hemp seed yield based on cultivation data. Bicakli et al. \cite{Bicakli2022-ip} demonstrated that random forest models can distinguish illegal cannabis crops from other vegetation in satellite imagery. In another study, Ferentinos et al. \cite{Ferentinos2019-io} introduced a deep learning system using transfer learning to identify diseases, pests, and deficiencies in cannabis plant images. Most recently, Boonsri et al. \cite{Boonsri2023-ob} applied deep learning-based object detection models to detect male and female cannabis seeds from augmented seed image datasets. Despite the progress of deep learning in many agricultural applications, its use for cannabis seed variety detection and classification has garnered less attention. Therefore, the main objective of this paper is to classify and detect seeds of 17 kinds of cannabis. 

The rest of the paper is organized as follows. Section 2 explores the related work. Section 3 describes the dataset, dataset preprocessing, training parameters, loss functions, Faster R-CNN architecture, and evaluation metrics. Section 4 presents the experimental results, discusses the findings, and compares the performance of the object detector. Finally, Section 5 summarizes the conclusions from our work.

\section{Related Work}
Seed identification has been tackled using both traditional methods \cite{Aznan2017-tp,Sharopova2002-zq}, image processing-based techniques \cite{Ahmad1999-st,Baek2020-hu}, and model-based approaches \cite{Thu_Hong2015-qj,Ali2020-tf}. Traditional methods include manual inspection and biochemical and molecular identification. The manual inspection methods are mainly based on the external shape (color, shape, size, etc.) of seeds \cite{Martin2010-zt}. However, it is difficult to classify seeds with similar external shapes \cite{Aznan2017-tp,Martin2010-zt}. The biochemical seed identification method can recognize seeds with different genetic characteristics, however identifying closely related varieties is a challenge. The molecular seed identification methods used DNA markers and they show stability and independence of environmental conditions in identifying seeds. However, their identification process can damage the sample and they are costly \cite{Sharopova2002-zq}. Image processing-based techniques include morphological operations, thresholding, segmentation, feature extraction, and texture analysis. Model-based approaches are divided into two categories: traditional machine learning models such as logistic regression, random forest, and decision tree. The second category encompasses deep learning models, including the commonly used convolutional neural networks \cite{islam_convolutional_2022}.

\subsection{Image processing in Seed Classification}
Traditional image processing techniques have been widely used for seed detection and analysis in previous research. For example, Ahmad et al. \cite{Ahmad1999-st} developed a model that distinguishes between asymptomatic and symptomatic soybean seeds based on color features. The methods used include RGB color analysis, thresholding for segmentation, discriminant analysis for feature selection, and linear discriminant classification with unequal priors to identify different seed damage types. Another study \cite{Baek2020-hu} proposed high-throughput methods using image analysis to efficiently measure the morphological and color traits of soybean seeds. The performance of traditional image processing pipelines depends heavily on extensive tuning and feature engineering specific to each seed type. They generalize poorly to unseen data and lack the learning capacity to improve with more training data.

\subsection{Machine learning in Seed Classification}
In recent years, there has been a lot of research carried out by researchers on the application of machine learning to the identification and classification of seeds and their other traits \cite{Zhao2022-xk}. Thu Hong et al. \cite{Thu_Hong2015-qj} introduced an automated classification system for distinguishing rice seed varieties, crucial for ensuring seed purity in rice production. The research employs various feature extraction techniques, including morphological, color, texture, GIST, and SIFT features, and evaluates the performance of machine learning classifiers such as KNN, SVM, and Random Forest. The results demonstrate that the Random Forest method, coupled with basic feature extraction techniques, achieves the highest classification accuracy of 90.54\%, for classifying the six rice varieties. In another research \cite{Ali2020-tf}, the authors trained Random Forest, BayesNet, LogitBoost, and Multilayer Perceptron to classify 6 corn seed varieties. Of all the classifiers, Multilayer Perceptron achieved the highest accuracy of 98.93\% on seed images. The effectiveness of supervised machine learning models relies significantly on feature extraction techniques. Inadequate or low-quality features can result in suboptimal performance.

\subsection{Deep Learning in Seed Classification}

Recent studies have begun leveraging deep learning for automated seed detection and classification. In one study \cite{Braguy2021-sn},  researchers used Faster R-CNN with various ResNet backbones to accurately detect, count, and discriminate germinated and non-germinated parasitic plant seeds from images. In a different study \cite{Heo2018-tc}, researchers introduced a real-time, high-throughput seed sorting system utilizing ResNet-18 image classifier and a batch inference strategy to achieve both high speed (500 fps) and high accuracy. This approach separated the object-detection task into localization and classification, demonstrating superior sorting accuracy (99.58\%) and purity (99.994\%) compared to commercial optical seed sorting systems. While deep learning has shown promising results for seed analysis, there remain considerable challenges. Collecting and annotating thousands of high-quality seed images needed to train a model is labor-intensive, time-consuming, and expensive.


Existing cannabis agriculture research has utilized deep learning to identify grow sites \cite{bicakli_cannabis_2022} and detect crop issues and pests \cite{ferentinos_image-based_2019}. However, the ability of the deep learning models to discriminate between cannabis seed varieties remains unexplored. Although deep learning approaches have shown promise for screening the gender of cannabis seeds \cite{Boonsri2023-ob}, the capability of these models to discriminate between seed varieties has yet to be explored. Our work seeks to bridge this research gap. In this paper, we use Faster R-CNN to accurately detect and delimit the bounding box region of the cannabis seed varieties. We evaluate the precision, recall, model complexity, and speed of the detector on the cannabis seed dataset to assess their suitability and performance for this agricultural application.

\section{Materials and Methods}
\subsection{Cannabis Seed Dataset}
\begin{figure}[t]
    \centering
    \includegraphics[width=0.4\textwidth]{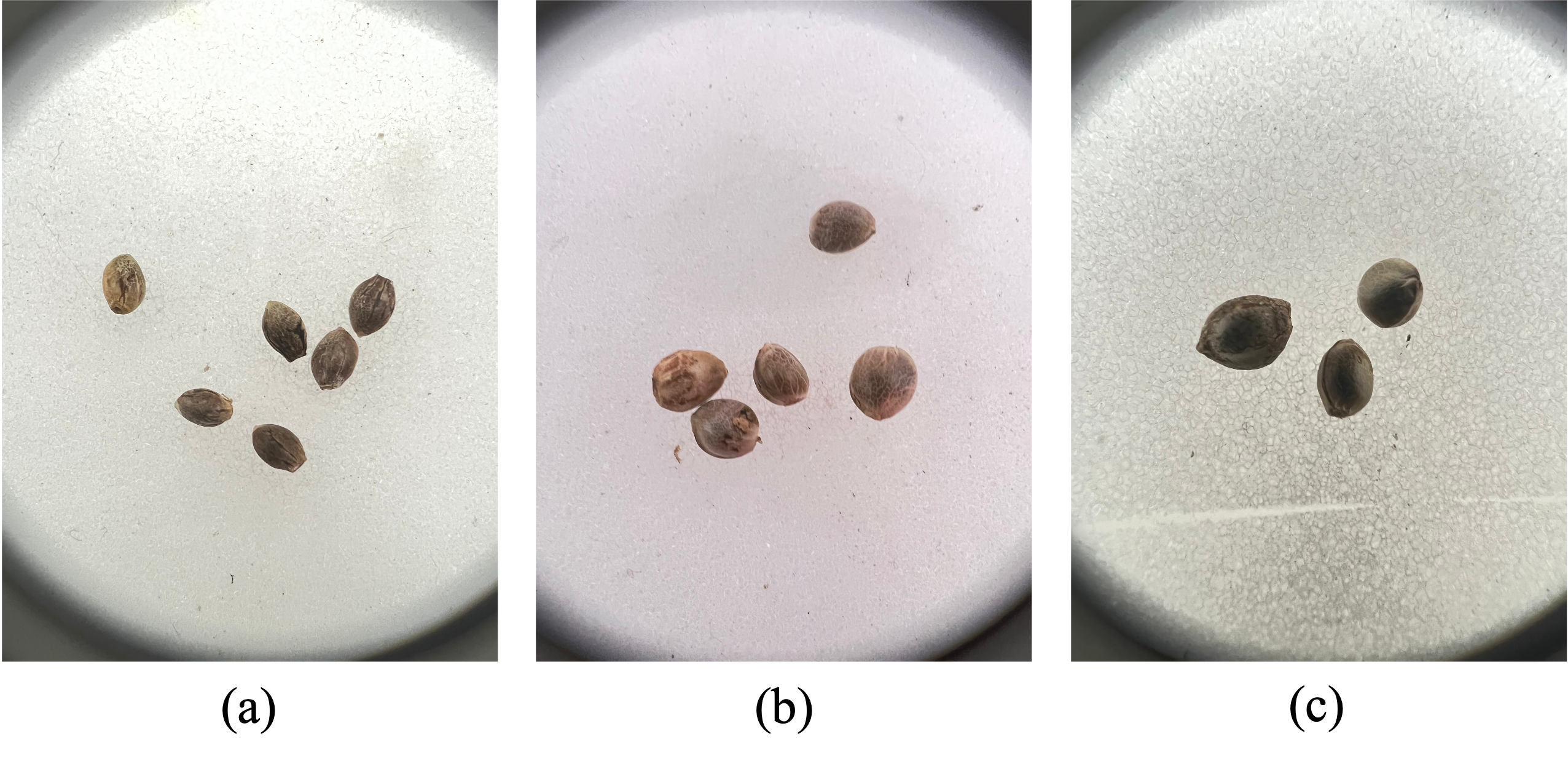}
    \vspace{-0.4cm}
    \caption{Three different cannabis seed types. (a) AK47, (b) Gelato, and (c) Gorilla Purple.}
    \label{fig:sample_seed}
    \vspace{-0.3cm}
\end{figure}

For this research, we selected a cannabis seed variety dataset \cite{Chumchu2023-yu} with 17 distinct classes. The seeds are locally available in Thailand. To the best of our knowledge, this is the first open-access dataset of cannabis seeds. The dataset contains 3335 high-resolution images of size 3023 x 4032 pixels, captured with an Apple iPhone 13 Pro. We remove 16 images due to their blurriness, leaving us with 3,319 images for this experiment. The images are taken with a consistent white background but with different lighting conditions and angles. Figure \ref{fig:sample_seed} shows the samples of different cannabis seed variants.

\subsection{Dataset Preprocessing}
Labeling each image manually is time-consuming and requires human labor. Instead of manual data annotation, we used Grounding DINO \cite{Liu2023-cj}, an open-set object detector, to extract bounding boxes from the dataset given an input text query. Open-set object detectors can detect arbitrary object categories rather than being limited to categories it was trained on. Grounding DINO can generate multiple 2D bounding boxes corresponding to an input image and noun phrases associated with a given image-text pair. This allowed us to extract object boundaries from the dataset without needing to manually draw boxes around each object instance.

After annotating the dataset, we split the dataset into training, validation, and test sets. We used 1771 images ($\sim$53\%) to train our object detection models and 723 images ($\sim$22\%) to validate the model during training time. The rest of the 825 images ($\sim$25\%) are held out to test the trained model's performance. 

Data augmentation is a technique to increase the dataset size when the data is limited. By applying various image transformations it helps the model to generalize on unseen data during the testing phase. We use Albumentations \cite{info11020125} image augmentation library to apply geometric transformations, color transformations, and blur operations on training images. The augmentation parameters are as follows: random flips with 50\% probability, maximum image shift of 0.0625, maximum scale change of 0.1, and maximum rotation angle of 45 degrees. The parameters also include randomly changing brightness and contrast within the range of 0.1 to 0.3 with a 20\% probability, random shifts to RGB channels by up to 10 intensity levels each, as well as maximum hue, saturation, and value shifts of 20, 30 and 20, respectively, for 10\% of the images. Additional augmentation parameters include a 10\% probability of applying channel shuffling and a 10\% probability of applying either random blur or median blur with a maximum kernel size of 3 during image augmentation.


%
\subsection{Training}
We use PyTorch \cite{NEURIPS2019_9015} and MMDetection \cite{mmdetection} to train Faster R-CNN models on an NVIDIA RTX 3090 GPU. We initialize Faster R-CNN with weights from detection models pre-trained on the COCO dataset. The models are then fine-tuned with the training dataset for 100 epochs with stochastic gradient descent optimization. A learning rate of 0.02 and 0.01 is used for Faster R-CNN along with a weight decay of 0.0001 and momentum of 0.9. We resize all images to a width of 360 and a height of 640 dimensions to serve as input to the models. The batch sizes for training, validation, and testing are configured as 2, 1, and 1, respectively.

\begin{table*}[htbp]
\caption{Mean Average Precision, Average Recall, F1 and Real-time Inference Results for Faster R-CNN}
\vspace{-0.4cm}
\label{map_ar}
\begin{center}
\begin{tabular}{|c|c|c|c|c|c|c|}
\hline
\textbf{Model} &  \textbf{mAP @ IoU:0.50:0.95} & \textbf{mAP @ IoU:0.50} & \textbf{Average Recall} & \textbf{F1} &  \textbf{Inference Speed (ms)} & \textbf{FPS}\\
\hline
mBaseline & 0.9168 & 0.9235 & 0.951 & 0.9336 & 55.6 & 18\\
mL1 &  \textbf{0.9408} & 0.9428 & \textbf{0.973} & \textbf{0.9566} & 59.5 & 16.8\\
mIoU & 0.9388 & \textbf{0.9431} & 0.969 & 0.9537 & 57.5 & 17.4\\
mGIoU & 0.9359 & 0.9396 & 0.964 & 0.9497 & 66.7 & 15\\
mDIoU & 0.938 & 0.9428 & 0.969 & 0.9532 & 61 & 16.4\\
mCIoU & 0.9365 & 0.9412 & 0.963 & 0.9496 & 57.8 & 17.3\\
\hline
\end{tabular}
\end{center}
\vspace{-0.5cm}
\end{table*}

\subsection{Faster R-CNN Network Architecture}
ResNet-50 \cite{He2015-kl} model pre-trained on ImageNet is used as the backbone of the Faster R-CNN network. The backbone is the initial processing layer in which features are extracted from the input image at several intermediate levels and act as the bottom-up pathway for the feature pyramid network. Feature pyramid network \cite{Lin2016-or} selects features generated from the last residual blocks from conv2, conv3, conv4, and conv4 of the ResNet 50 backbone to define the pyramid level. Our FPN has four lateral 1x1 convolutions, which merge with the upsampled top-down pathway and reduce the four output feature maps [256, 512, 1024, 2048] from the bottom-up pyramid levels to fixed 256-channel outputs. We apply four 3x3 convolutions, each taking the merged feature map as input and outputting a 4-level feature pyramid, with each level having a 256-channel feature map. Feature maps from the FPN are then forwarded to the Region Proposal Network (RPN) \cite{ren_faster_2017}. RPN operates by sliding a small network on the feature maps extracted from the feature pyramid. Our RPN head utilizes a 3x3 convolution with 256 channels followed by ReLU activation to process the 256-channel feature map from the convolutional layer as input. For each sliding window, our anchor generator uses one scale (8) and three aspect ratios (1:2, 1:1, 2:1) to generate regional proposals. Each region proposal is then passed to a 1x1 256-channel convolution layer for classification and another parallel 1x1 12-channel convolution layer to regress the bounding box coordinates. We apply non-maximum suppression (NMS) based on their classification scores to reduce redundancy among region proposals. We set the IoU threshold for NMS to 0.7, retaining a maximum of 1000 region proposals per image and removing boxes with high overlap. 

The region proposals are then fed into the Fast R-CNN branch that extracts features using 4 ROI Align \cite{He2017-el} layers to output a fixed size 7x7 features for each Proposal. The feature vector is fed into two fully connected layers: a softmax layer to predict the class score with cross-entropy loss and regression layers to predict the bounding box location. If the IoU is greater than the IoU threshold value of 0.5, the proposal is identified as a positive result. In the bounding box regressor layer, we used five different loss functions, including L1 loss, IoU loss, Generalized IoU (GIoU) loss \cite{Rezatofighi2019-tc}, Distance IoU (DIoU) loss \cite{Zheng2019-pj}, and Complete IoU (CIoU) loss \cite{Zheng2020-gj} loss to minimize the loss for the regression layer.

\subsection{Loss Functions}
The five loss functions we use in the bounding box regression layer of Faster R-CNN are defined as follows:
\subsubsection{L1 loss}
L1 loss calculates the absolute difference between the coordinates of the ground truth bounding box and the predicted bounding box. It is defined as,

\vspace{-0.2cm}
\begin{equation}
\label{eq:l1loss}
    L_{L 1}=\frac{\sum_{i=1}^n\left|B^G-B^P\right|}{n}  
\end{equation}

From \eqref{eq:l1loss}, L1 loss is the ratio of the sum of absolute differences between the ground truth bounding box $B^G$ and the predicted bounding box $B^P$ divided by the total number of elements n. The range of $L_{1}$ loss is $0 \le L_{L1} \le \infty$.
\subsubsection{IoU loss}
Intersection over Union (IoU) is a metric that compares the similarity between two arbitrary shapes. The IoU computation for the ground truth bounding box $B^G$ and predicted bounding box $B^P$ is defined as

\vspace{-0.2cm}
\begin{equation}
\label{eq:iou}
    IoU=\frac{\left|B^G \cap B^P\right|}{\left|B^G \cup B^P\right|}
\end{equation}

It is the area of the intersection of $B^G$ and $B^P$ divided by the union area of $B^G$ and $B^P$ defined in \eqref{eq:iou}. Therefore, \eqref{eq:iouloss} defines the IoU loss as

\vspace{-0.4cm}
\begin{equation}
\label{eq:iouloss}
    L_{IoU}= 1 - IoU
\end{equation}

The range of $IoU$ is $0 \le IoU \le 1$, and $L_{IoU}$ is  $0 \le L_{IoU} \le 1$.

\subsubsection{GIoU loss}
IoU loss can not tell if the two bounding boxes are close or far from each other if there is no intersection between the ground truth and predicted bounding boxes. In non-overlapping cases, the gradient of IoU loss becomes zero. However, the GIoU loss \cite{Rezatofighi2019-tc} addresses this issue of vanishing gradient for non-overlapping cases by introducing a penalty term that considers the smallest enclosing rectangle of both bounding boxes and moves the predicted box toward the target box. The equation for GIoU loss is

\vspace{-0.5cm}
\begin{equation}
\label{eq:giouloss}
    L_{G I o U}=1-G I o U=1-I o U+\frac{A^C \backslash\left(B^G \cup B^P\right)}{A^C}
\end{equation}

$A^{C}$ is the area of the smallest enclosing rectangles of the two bounding boxes in \eqref{eq:giouloss}, and it acts as a penalty term to move the predicted bounding box closer to the ground truth bounding box. The range of GIoU is $-1 \le GIoU(B^G, B^P) \le 1$, and the range of $L_{GIoU}$ is $0 \le L_{GIoU} \le  2$. Though GIoU solves the problem of vanishing gradient of non-overlapping boxes, it has problems of slow convergence and inaccurate bounding box regression.
\subsubsection{DIoU loss}
DIoU loss \cite{Zheng2019-pj} minimizes the distance between the center points of the ground truth bounding box and the predicted bounding box. It is defined as

\vspace{-0.2cm}
\begin{equation}
\label{eq:diouloss}
    L_{D I o U}=1-I o U+\frac{\rho^2\left(b^p, b^g\right)}{c^2}
\end{equation}

From \eqref{eq:diouloss}, $b^p$ and $b^t$ are the central points of the ground truth and predicted bounding box, and c is the diagonal distance of the smallest box covering the two boxes. $\rho^2(b^p, b^g)$ is the Euclidean distance between two center points of $b^p$ and $b^g$. DIoU loss minimizes the distance between the central point of the ground truth bounding box and the predicted bounding box, and it converges faster than GIoU loss.
\subsubsection{CIoU loss}
CIoU loss \cite{Zheng2020-gj} considers the overlap area, central point distance, and aspect ratio for regressing the bounding box coordinates. CIoU loss is based on DIoU loss and adds an aspect ratio parameter defined as

\vspace{-0.2cm}
\begin{equation}
\label{eq:ciouloss}
    L_{D I o U}=1-I o U+\frac{\rho^2\left(b^p, b^g\right)}{c^2}+\alpha V
\end{equation}

In \eqref{eq:ciouloss}, $V$ measures the consistency of the aspect ratio, and $\alpha$ is the trade-off parameter. When IoU between predicted and ground truth boxes is less than 0.5, CIoU loss reverts to DIoU loss. This behavior reflects that the consistency of the aspect ratio is not essential when there is minimal overlap between two boxes. Conversely, when IoU is greater or equal to 0.5, the consistency of the aspect ratio becomes critical, and CIoU loss refines the localization accuracy through the aspect ratio parameters $\alpha V$.



%

\begin{table*}[t]
\caption{Classwise mAP at IoU threshold $.50 \le IoU \le .95$ and at IoU threshold $.50$}
\vspace{-0.4cm}
\label{classwise_map}
\begin{center}
\begin{tabular}{|c|c|c|c|c|c|c|c|c|c|c|c|c|}
\hline
&\multicolumn{6}{|c|}{\textbf{mAP @ IoU:0.50:0.95}} &\multicolumn{6}{|c|}{\textbf{mAP @ IoU:0.50}}\\
\cline{2-13} 
\textbf{Class} & \textbf{mBasline} & \textbf{mL1} & \textbf{mIoU} & \textbf{mGIoU} & \textbf{mDIoU} & \textbf{mCIoU} & \textbf{mBasline} & \textbf{mL1} & \textbf{mIoU} & \textbf{mGIoU} & \textbf{mDIoU} & \textbf{mCIoU}\\
\hline
AK47 & 0.984 & 0.984 & 0.988 & \bfseries{0.989} & 0.987 &0.986 & 0.985 & 0.984 & 0.988 & \bfseries{0.989} & 0.988 & 0.987\\
BBA & 0.992 & 0.99 & \bfseries{0.993} & 0.991 & 0.992 & 0.99 & \bfseries{0.998} & 0.995 & \bfseries{0.998} & \bfseries{0.998} & \bfseries{0.998} & \bfseries{0.998}\\
CP & 0.291 & \bfseries{0.399} & 0.366 & 0.337 & 0.377 & 0.352 & 0.296 & \bfseries{0.403} & 0.369 & 0.343 & 0.387 & 0.355\\
GELP & 0.951 & 0.956 & \bfseries{0.966} & 0.957 & 0.963 & 0.958 & 0.957 & 0.956 & \bfseries{0.966} & 0.957 & 0.964 & 0.959\\
GP & 0.926 & 0.972 & 0.975 & 0.971 & \bfseries{0.976} & 0.971 & 0.946 & 0.978 & 0.982 & 0.978 & \bfseries{0.984} & 0.981\\
HKRKU & 0.998 & \bfseries{1} & \bfseries{1} & 0.998 & 0.999 & \bfseries{1} & \bfseries{1} & \bfseries{1} & \bfseries{1} & 0.999 & \bfseries{1} & \bfseries{1}\\
HKRPPST1 & 0.934 & 0.957 & 0.954 & 0.954 & \bfseries{0.959} & 0.952 & 0.949 & 0.959 & 0.959 & 0.956 & \bfseries{0.965} & 0.959\\
HSSNTT1 & 0.947 & 0.949 & 0.967 & \bfseries{0.976} & 0.968 & 0.967 & 0.949 & 0.949 & 0.968 & \bfseries{0.977} & 0.969 & 0.967\\
KDKT & 0.952 & \bfseries{0.975} & 0.97 & 0.964 & 0.966 & 0.971 & 0.953 & \bfseries{0.975} & 0.972 & 0.964 & 0.966 & 0.971\\
KD & 0.925 & \bfseries{1} & 0.987 & \bfseries{1} & 0.987 & 0.985 & 0.933 & \bfseries{1} & \bfseries{1} & \bfseries{1} & \bfseries{1} & \bfseries{1}\\
KKV & 0.953 & \bfseries{0.995} & 0.989 & 0.986 & 0.988 & 0.987 & 0.954 & \bfseries{0.999} & 0.994 & 0.994 & 0.994 & 0.993\\
PD & 0.995 & \bfseries{0.999} & 0.995 & 0.997 & 0.996 & 0.996 & \bfseries{1} & \bfseries{1} & \bfseries{1} & \bfseries{1} & \bfseries{1} & \bfseries{1}\\
SDA & 0.999 & \bfseries{1} & \bfseries{1} & 0.998 & 0.999 & \bfseries{1} & \bfseries{1} & \bfseries{1} & \bfseries{1} & \bfseries{1} & \bfseries{1} & \bfseries{1}\\
SKA & 0.973 & 0.976 & 0.973 & 0.975 & 0.974 & \bfseries{0.984} & 0.979 & 0.977 & 0.976 & 0.976 & 0.977 & \bfseries{0.987}\\
TFT & 0.987 & \bfseries{0.996} & 0.995 & \bfseries{0.996} & \bfseries{0.996} & 0.995 & \bfseries{1} & \bfseries{1} & \bfseries{1} & \bfseries{1} & \bfseries{1} & \bfseries{1}\\
TKDRD1 & 0.841 & 0.869 & \bfseries{0.888} & 0.867 & 0.866 & 0.863 & 0.847 & 0.873 & \bfseries{0.895} & 0.877 & 0.871 & 0.871\\
TKKWA1 & 0.938 & \bfseries{0.976} & 0.953 & 0.955 & 0.953 & 0.964 & 0.954 & \bfseries{0.98} & 0.965 & 0.966 & 0.965 & 0.972\\
\hline
\end{tabular}
\end{center}
\vspace{-0.5cm}
\end{table*}

\subsection{Evaluation Metrics}
We use mean average precision (mAP), recall, and F1 \cite{ahmed_dsteelnet_2023} evaluation metrics to evaluate the performance change in our object detection models. mAP is the most commonly used metric to assess the performance of our object detection models. It has three parameters: IoU, precision, and recall. We evaluate our models using a range of IoU thresholds, from 0.50 to 0.95 in 0.05 increments. As the IoU threshold increases, the criterion for considering a predicted bounding box as a true positive becomes stricter and requires a high overlap for a detection to be considered positive. For each IoU threshold, we calculate precision and recall. Precision is the ratio of correctly predicted true positives to the total predicted positives. True Positives are the seeds with an IoU value greater than a given threshold, with correctly identified class labels. Consequently, False positives are seeds with an IoU less than the IoU threshold, or the class label of the predicted bounding box is incorrect. Recall is the ratio between the true positive and the total positive samples. Here, false negatives are the seeds that are incorrectly predicted as negatives by the model. Specifically, mAP is calculated by computing the average Precision for individual seed classes and averaging the average precision across all classes. Additionally, we report F1 scores. F1 is the harmonic mean of precision and recall. The harmonic mean gives more weight to small values, which means if the recall or precision score is low, it will lower the F1 score.

\section{Experimental Evaluation}
This section will present a detailed description of the Faster R-CNN model's performance on the evaluation metrics. 
We evaluate our experiments with the six different Faster R-CNN models we trained, 1) a baseline model trained with no image augmentation and L1 loss (mBaseline), 2) with image augmentation and L1 loss (mL1), 3) with image augmentation and IoU loss (mIoU), 4) with image augmentation and Generalized IoU loss (mGIoU), 5) with image augmentation and Distance IoU loss (mDIoU), and 6) with image augmentation and Complete IoU loss (mCIoU). 


Table~\ref{map_ar} gives a comparison of the six Faster R-CNN models, evaluating their performance across various key metrics. The metrics include mAP at different IoU thresholds, specifically from 0.50 to 0.95 in increments of 0.05, mAP at IoU:0.50, Average Recall, and F1 score. Our mBaseline model serves as a reference point for evaluating the performance of the other models. The mL1 model demonstrates the highest mAP at IoU:0.50:0.95, with values of 0.9408. Additionally, the mL1 model achieves the highest average recall of 0.973. The mIoU, mGIoU, mDIoU, and mCIoU models exhibit relatively similar performance levels at IoU:0.50. However, the mIoU model demonstrates the highest mAP at IoU threshold 0.50 with a value of 0.9431. The mL1 model achieves the top F1 score of 0.9566, exhibiting strong precision-recall balance. The mIoU model follows closely with a 0.9537 score, while the mBaseline model, without any image augmentation applied during the training, lags significantly at 0.9336 F1 measure. Overall, the mL1 model stands out as the most effective among the listed models, offering the highest mAP, recall, and F1 values, which suggests that mL1 excels in capturing objects with varying levels of spatial overlap.

The classwise mAP@0.50:0.95 and mAP@0.50 of cannabis seed variants are shown in Table~\ref{classwise_map}. In terms of the broader IoU range (0.50:0.95), all loss functions perform quite similarly, with mL1 and mIoU slightly outperforming the mGIoU, mDIoU, and mCIoU on average across classes. However, mL1 overall achieves the best performance, with the highest mAP scores for 9 out of the 17 classes. When looking at mAP specifically for an IoU threshold of 0.50, no single loss function dominates the performance across classes. The analysis of the provided tables reveals an interesting pattern regarding the L1 Loss function, which consistently yields the highest mAP scores across nine out of the seventeen classes in both IoU threshold scenarios. Moreover, `CP'  exhibits the biggest mAP boost from using L1 loss compared to the mBaseline, increasing mAP by 37.11\% at high IoU and by 36.15\% at IoU 0.50. This suggests that L1 regression loss provides advantages for improving localization of this more challenging class.  


\begin{figure}[h]
    \centering
    \includegraphics[width=0.4\textwidth]{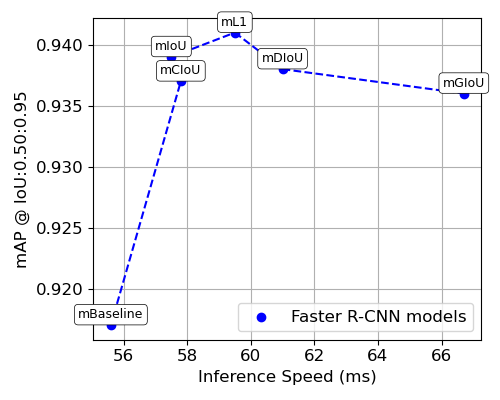}
    \caption{mAP vs Inference Speed of Faster R-CNN Models}
    \label{fig:map_inference}
\end{figure}

We see a tradeoff between accuracy and speed while comparing the mAP and inference speeds of the Faster R-CNN models in Table \ref{map_ar}. The best-performing mL1 lags in inference speed by 6.67\% compared to the fastest mBaseline model.  However, Figure \ref{fig:map_inference} depicts that mIoU strikes the best balance by achieving the second-highest mAP (0.9388) after mL1 and an inference speed of 57.5 ms, which is faster than the top-performing mL1 model. The small compromise in speed in mIoU compared to the fastest mBaseline is justified by sizable gains in detection precision. Therefore, mIoU provides the best tradeoff when both accuracy and real-time performance need to be optimized.


\section{Conclusion}
This experimental evaluation of Faster R-CNN models for cannabis seed variant detection unveils nuanced insights into their performance across key metrics. The mL1 model emerges as the most effective, showcasing superior mAP at various IoU thresholds, with a peak value of 0.9408. Notably, mL1 also excels in average recall and achieves the highest F1 score of 0.9566, indicating a commendable balance between precision and recall. While advanced loss functions like Generalized IoU (mGIoU), Distance IoU (mDIoU), and Complete IoU (mCIoU) offer marginal improvements, the L1 loss consistently yields the highest mAP scores across a majority of seed classes, revealing its robust performance across evaluation scenarios. Real-time inference performance reveals a trade-off between accuracy and speed, with the mBaseline model being the fastest but less accurate, while mGIoU sacrifices speed for enhanced precision. This analysis provides valuable insights for practitioners, emphasizing the importance of selecting an appropriate loss function based on specific priorities such as accuracy requirements and real-time processing constraints in cannabis seed variant detection applications. In the future, we aim to expand this research and implement transformer model for detecting cannabis seed variants on a more robust dataset.



\end{document}